\documentclass[sigconf,natbib=true,svgnames]{acmart}

\usepackage[dvipsnames,svgnames,x11names]{colortbl}
\usepackage{stfloats}
\usepackage{multirow}
\usepackage{booktabs}
\usepackage{amsthm,amsmath}
\usepackage{mathrsfs}
\usepackage{graphicx}
\usepackage{subfigure}
\usepackage{makecell}
\usepackage{pifont}

\AtBeginDocument{%
  \providecommand\BibTeX{{%
    \normalfont B\kern-0.5em{\scshape i\kern-0.25em b}\kern-0.8em\TeX}}}

\copyrightyear{2023}
\acmYear{2023}
\setcopyright{acmlicensed}
\acmConference[SIGIR '23]{Proceedings of the 46th International ACM SIGIR Conference on Research and Development in Information Retrieval}{July 23--27, 2023}{Taipei, Taiwan}
\acmBooktitle{Proceedings of the 46th International ACM SIGIR Conference on Research and Development in Information Retrieval (SIGIR '23), July 23--27, 2023, Taipei, Taiwan}
\acmPrice{15.00}
\acmDOI{10.1145/3539618.3591758}
\acmISBN{978-1-4503-9408-6/23/07}

\begin{document}

\title{Rethinking Benchmarks for Cross-modal Image-text Retrieval}


\author{Weijing Chen}
\email{chenweijing17@ruc.edu.cn}
\affiliation{%
  \institution{Renmin University of China}
  \city{Beijing}
  \country{China}
}

\author{Linli Yao}
\email{linliyao@ruc.edu.cn}
\affiliation{%
  \institution{Renmin University of China}
  \city{Beijing}
  \country{China}
}

\author{Qin Jin}
\authornote{Corresponding Author.}
\email{qjin@ruc.edu.cn}
\affiliation{%
  \institution{Renmin University of China}
  \city{Beijing}
  \country{China}
}

\newcommand{\qin}[1]{{\color{red}{\bf Qin}: {#1}}}



\begin{abstract}
Image-text retrieval, as a fundamental and important branch of information retrieval, has attracted extensive research attentions. The main challenge of this task is cross-modal semantic understanding and matching. Some recent works focus more on fine-grained cross-modal semantic matching. With the prevalence of large scale multimodal pretraining models, several state-of-the-art models (\emph{e.g.} X-VLM) have achieved near-perfect performance on widely-used image-text retrieval benchmarks, \emph{i.e.} MSCOCO-Test-5K and Flickr30K-Test-1K.
In this paper, we review the two common benchmarks and observe that they are insufficient to assess the true capability of models on \emph{fine-grained} cross-modal semantic matching. The reason is that a large amount of images and texts in the benchmarks are coarse-grained. 
Based on the observation, we renovate the coarse-grained images and texts in the old benchmarks and establish the improved benchmarks called \emph{MSCOCO-FG} and \emph{Flickr30K-FG}. Specifically, on the image side, we enlarge the original image pool by adopting more similar images. On the text side, we propose a novel semi-automatic renovation approach to refine coarse-grained sentences into finer-grained ones with little human effort. 
Furthermore, we evaluate representative image-text retrieval models on our new benchmarks to demonstrate the effectiveness of our method. We also analyze the capability of models on fine-grained semantic comprehension through extensive experiments. The results show that even the state-of-the-art models have much room for improvement in fine-grained semantic understanding, especially in distinguishing attributes of close objects in images. Our code and improved benchmark datasets are publicly available\footnote{\url{https://github.com/cwj1412/MSCOCO-Flikcr30K_FG}}, which we hope will inspire further in-depth research on cross-modal retrieval. 
\end{abstract}





\settopmatter{printfolios=true}

\maketitle

\section{Introduction}

\begin{figure}[!t]
  \centering
  \includegraphics[width=0.75\linewidth]{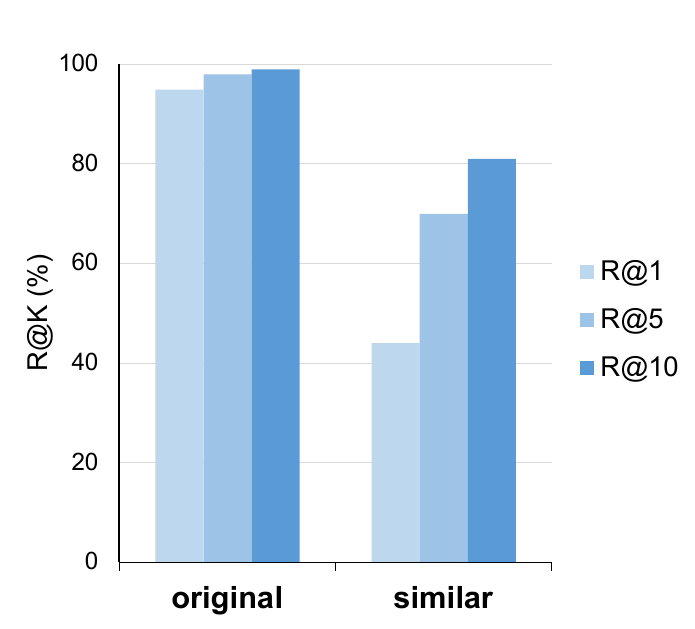}
  \vspace{-8pt}
  \caption{The performance of state-of-the-art model X-VLM in text-to-image (T2I) retrieval on different candidate images. We observe a significant performance degradation when retrieving on similar images.}
  \label{fig1}
\end{figure}

Image-text retrieval, as one of the basic topics of cross-modal research, has a wide range of applications in real-world scenarios such as search engines, recommendation systems, and question answering systems. It requires the machine to retrieve images given text queries or find matching descriptions given image queries. The key challenge in image-text retrieval is learning semantic correspondence across different modalities. To achieve better cross-modal retrieval, current mainstream image-text retrieval works tend to focus more on fine-grained cross-modal semantic comprehension, where various well-deigned alignment and reasoning modules, including attention mechanisms \cite{scan,bfan,pfan,saem,camp}, graph-based networks \cite{vsrn,gsmn,sgraf}, and scene graphs \cite{scg,sgm} etc. are adopted.

The introduction of large scale multimodal pretraining models~\cite{clip,uniter,xvlm} in recent years has brought significant performance gains in image-text retrieval. Great efforts have been made to design pretraining tasks for some existing pretraining models to stimulate fine-grained semantic understanding, leading to near-perfect performances in image-text retrieval. Several state-of-the-art models, such as X-VLM \cite{xvlm}, have already achieved 98.2 and 90.5 on R@10 for image-to-text (I2T) retrieval and text-to-image (T2I) retrieval on one of the common benchmarks MSCOCO-Test-5K. 

The near-perfect performance of current models is exciting, but also arouses our curiosity, that is, do these models really have human-comparable cross-modal semantic knowledge, especially in fine-grained semantic understanding? Or are existing benchmarks insufficient to validate the real model abilities in cross-modal fine-grained semantics understanding?
We therefore perform a mini text-to-image (T2I) retrieval experiment to test our hypothesis. 
Specifically, we create two types of test settings. First, we randomly select 100 image-text pairs from MSCOCO-Test-5K as our mini-test(original) setting. Then, for a given text query in mini-test(original), we deliberately select 99 images similar to its corresponding target image from MSCOCO-Test-5K and auxiliary sources (see Sec. \ref{section:image}), which together with the target image are used as the more challenging candidate set for T2I retrieval, which is named as our mini-test(similar) setting. The performance of X-VLM under these two mini experiment settings\footnote{The performance under the mini-test(similar) setting is averaged over all 100 queries.} is shown in Figure \ref{fig1}, where a significant performance degradation can be observed from mini-test(original) to mini-test(similar). These observations inspire us to review current benchmarks, and we discover two problems. 

\begin{figure*}[t]
  \centering
  \includegraphics[width=0.9\linewidth]{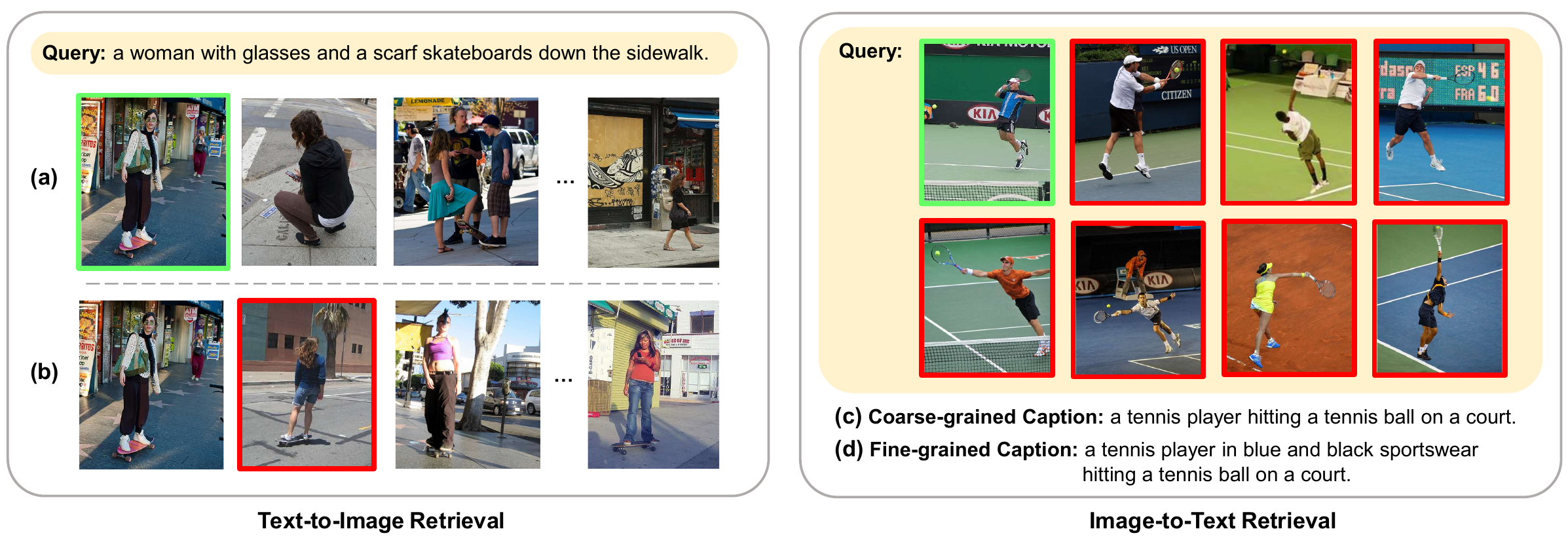}
  \vspace{-8pt}
  \caption{Illustration of 'coarse-grained images and texts' issues in current benchmarks: (a) candidate images vary widely which leads the model to easily match the target image without fine-grained semantic understanding; (b) however, the model fails without decent fine-grained semantic understanding if candidate images are similar; (c) coarse-grained text leads to unexpected match with non-target image queries in the benchmark; (d) fine-grained text can well distinguish similar image queries.}
  \label{fig2}
\end{figure*}

First, \textbf{the images used for benchmarking are 'coarse-grained'}, which refers to that the sizes of the image retrieval pools are small, and images in the pools vary widely (or the pool is semantically sparse), resulting in easily distinguishable retrieval targets without requiring fine-grained semantic understanding. The two common image-text retrieval benchmarks, MSCOCO-Test-5K \cite{mscoco} and Flickr30K-Test-1K \cite{flickr30k}, contain only 5000 and 1000 images respectively. Moreover, these images are randomly selected from Flickr30K and MSCOCO. The limited scale and semantically sparse retrieval candidate pool can lead to systematic weakness for the image-text retrieval evaluation. 
As the example shown in Figure \ref{fig2}(a), for the given query "\textit{a woman with glasses and a scarf skateboards down the sidewalk}", the retrieval pool contains a wide variety of images, most of which are semantically distinct from the query. In such case, the model can easily pick the target image from the candidate image pool without fine-grained semantic understanding. However, if we require the model to retrieve from similar candidate images, the model may fail without decent fine-grained semantic understanding ability, as shown in Figure \ref{fig2}(b).

Second, \textbf{the texts used for benchmarking are 'coarse-grained'}. A large number of text descriptions in the benchmark are not detailed enough. In MSCOCO and Flickr30K, five sentences are manually annotated for each image, some are finer-grained (\emph{e.g.} \textit{"a giraffe stands in a green field with a house in the background"}), while others are coarse-grained (\emph{e.g.} \textit{"a giraffe stands in a field"}). 
Nevertheless, the coarse-grained texts may cause issues in the image-text retrieval evaluation, as different image queries may well match the same coarse-grained text, as shown in Figure \ref{fig2}(c). Such an evaluation case is not a valid test sample for validating model capabilities. However, if the text becomes finer-grained, it can be a good sample to verify the fine-grained semantic understanding capability of models, as shown in Figure \ref{fig2}(d).

To address the above described deficiencies in current benchmarks, we propose to renovate existing benchmarks and build corresponding new benchmarks, namely \textbf{MSCOCO-FG} and \textbf{Flickr30K-FG}. 
We first adopt similar images to enlarge the original candidate pool to obtain a larger and semantically denser pool. Concretely, we leverage a multimodal model to search images similar to the target images from the original pool and auxiliary sources, and then combine all similar images including the target images to form a new image pool.
Moreover, we propose a novel semi-automatic method to refine the coarse-grained texts. Specifically, vision-language pretraining models are leveraged to detect potential coarse-grained texts first, which are then enriched via prompting new details. Finally, some manual corrections are performed on the refined texts. 

We further carry out extensive experiments to test various existing models on our new benchmarks. The experiment results show that models that have achieved near-perfect performance in the old benchmarks no longer perform perfectly in our new benchmarks, indicating that current image-text retrieval models still have much room for improvement in cross-modal fine-grained semantic understanding.
The main contributions of this work include:
\begin{itemize}
\setlength{\itemsep}{0pt}
\setlength{\parsep}{0pt}
\setlength{\parskip}{0pt}
\item we review the current image-text retrieval benchmarks and identify that they are insufficient to assess the fine-grained semantic understanding ability of models.

\item we renovate the old benchmarks and build the new fine-grained benchmarks MSCOCO-FG and Flickr30K-FG for image-text retrieval, through exploiting similar images to expand image pools and improving coarse-grained texts semi-automatically. The new benchmarks will be released to support more in-depth research.

\item we evaluate several representative image-text retrieval models on our new benchmarks to prove the validity of our method, and further analyze their capabilities in comprehending cross-modal fine-grained semantics.
\end{itemize}

\section{Related work}
\subsection{Image-Text Retrieval}
Various methods have been proposed for image-text retrieval over the years, which can be broadly categorized into two groups.

\textbf{Non-pretraining models.}
Earlier works \cite{devise,dan,vse++} like VSE++ \cite{vse++} mainly use Convolutional Neural Networks (CNN) to extract fixed grid features to represent images, which only provide local pixel-level information and suffer from comprehending high-level semantic concepts. To solve this problem, following works \cite{scan,bfan,pfan,saem} apply an object detector to better encode images. For example, SCAN \cite{scan} adopts Faster R-CNN \cite{rcnn} to detect objects in images and align them to words of sentences. More recently, researchers explore diverse approaches \cite{camp,vsrn,scg,sgm,camera,sgraf,dime,naaf} for finer-grained semantic alignments between images and texts. Particularly, VSRN \cite{vsrn} exploits Graph Convolutional Networks (GCN) \cite{gcn} and Gated Recurrent Unit (GRU) \cite{gru} to perform local-global semantic reasoning. SGM \cite{sgm} constructs external scene graphs \cite{scenegraph} to enhance learning of visual relationships. DIME \cite{dime} is proposed to dynamically learn modality interaction patterns.

\textbf{Vision-language pretraining models.}
Vision-language pretraining (VLP) aims to learn vision language alignments from a large amount of image-text pairs through self-supervised tasks. After fine-tuning on the downstream image-text retrieval task, VLP models outperform non-pretraining models by a significant margin. Their architectures can be roughly divided into two types: one-stream and two-stream. 
Models with one-stream structure \cite{unicodervl,uniter,oscar,vinvl,villa,vilt} encode image and text with the same encoder, which learns vision language alignments through different pretraining tasks. 
For example, UNITER \cite{uniter} applies a unified Transformer encoder to learn contextualized embedding of image regions and words in a common space.
Models with two-stream structure \cite{vilbert,ernievil,clip,soho,unimo,meter,xvlm} first encode image and text separately with independent encoders and interact cross-modal semantics with co-attention layers, providing more flexible encoding of image and text. 
Specifically, CLIP \cite{clip} adopts an image encoder using ResNet-50 \cite{resnet50}/ViT \cite{vit} and encodes text with a Transformer.
After contrastive pretraining on 400 million image-text pairs, the separate encoders gain abilities to encode universal representations for two modalities.

In order to investigate the capabilities of previous models in fine-grained cross-modal semantic comprehension and alignment, we select several representative non-pretraining and pretraining models to evaluate on our new benchmarks and provide detailed analysis in the following experiments.

\subsection{Image-Text Datasets}
With the increasing interest in cross-modal research, various cross-modal datasets, especially image-text datasets, have been proposed to support in-depth exploration. Existing image-text datasets can be divided into two categories according to the data collection methods. 

\textbf{Datasets with human annotations.}
Early on, researchers create image-text datasets by first crawling images from the Internet, and then manually annotating labels, regions and textual descriptions for the images.
Following \citet{flickr8k}, \citet{flickr30k} introduce Flickr30K with 31,783 images of everyday activities and scenes collected from the Flickr website and 5 annotated descriptions for each image via crowd-sourcing (Amazon Mechanical Turk). \citet{mscoco} propose a larger and more detailed dataset named MSCOCO with 123,287 images of 91 common object categories (also collected from Flickr) and manually annotate bounding boxes, segmentations and 5 descriptions for each image. Later, \citet{vg} propose Visual Genome based on MSCOCO with additional descriptions for visual regions to provide richer details of images. 
Due to their high quality, these datasets are widely used in training and evaluating various image-text tasks. 

\textbf{Datasets with filtered web resources.} 
As the vision-language pretraining models perform remarkably in the natural language processing field, researchers tend to increase the size of image-text datasets for training better multimodal models. To overcome the limitations of manually annotated datasets, they mostly construct datasets with web resources without human annotations. \citet{yfcc} harvest about 99 million images with their metadata including tags, locations from Flickr to construct YFCC-100M. Since the web data are often noisy, SBU Captioned Photo Dataset \cite{sbu} and Conceptual Captions \cite{cc3m} extract the alt-text of the original HTML text and clean them following several predefined rules to obtain more accurate and detailed image descriptions. Besides relying on alt-text, some datasets employ other text sources. For example, RedCaps \cite{redcaps} collects captions written by social media users to construct image-text pairs. 
Although these datasets are quantitatively superior to the annotated datasets, their quality is flawed even with well-designed filtering pipelines. Therefore, they are mostly used in the pretraining stage and are not suitable for evaluation.

To sum up, image-text datasets with human annotations are more appropriate for benchmarking, and therefore common image-text retrieval benchmarks are mainly based on manually annotated datasets, e.g MSCOCO and Flickr30K. However, these benchmarks are still not sufficient for fine-grained image-text retrieval assessment. We then propose to renovate current benchmarks to build new benchmarks via semi-automatic approaches, enabling fine-grained evaluations and saving the cost of manual re-annotation.


\section{Method}
\label{section:method}
A conventional image-text retrieval benchmark $\mathcal{B}$ normally contains $N$ images and each one has $M$ corresponding textual descriptions, and it is usually used to assess two sub-tasks: \textit{Text-to-Image (T2I)} retrieval and \textit{Image-to-Text (I2T)} retrieval. Text-to-image retrieval aims to search the target image from the whole candidate image pool $\mathcal{I}$ given the text query, while image-to-text retrieval requires the model to search at least one target text description from the candidate text pool $\mathcal{D}$ given the image query. 
As we observe, current benchmarks suffer from coarse-grained images and texts, making them incapable to provide fine-grained image-text retrieval evaluations. Considering that building a completely new benchmark from scratch is very expensive and time-consuming, we propose to renovate the existing benchmarks $\mathcal{B}$ to meet the higher fine-grained evaluation requirements. 
The following Section \ref{section:image} first presents how we build the more challenging and fine-grained image pools. Section \ref{section:text} describes our semi-automatic approaches for refining texts from coarse-grained to fine-grained.

\subsection{Renovate image candidate pool}
\label{section:image}
Since the images in current image retrieval pools are widely distinct, models can easily select the correct target image given the text query without fine-grained understanding. To improve existing benchmarks to support fine-grained semantic evaluation, we leverage similar images to enlarge image pools.
Denoting the original image pool as $\mathcal{I_{\textrm{old}}}$ and each image (target image) in the pool as $I_t$ (t$\in$[0,N]), we search similar images for $I_t$ to expand $\mathcal{I_{\textrm{old}}}$ to the new pool $\mathcal{I_{\textrm{new}}}$ through three steps: 
(i) prepare a non-overlap image pool $\mathcal{I_{\textrm{can}}}$ to provide candidate images, 
(ii) search a similar image set $S_t$ for each $I_t$ from $\mathcal{I_{\textrm{can}}}$,
and (iii) gather all similar image sets $\{S_t\}$ into a new image pool $\mathcal{I_{\textrm{new}}}$.

\subsubsection{\textbf{Prepare candidate images.}}
Before searching similar images to expand the original image pool, we should first determine which images can be candidates for similar images. In addition to the images in the original pool, we need more candidate images. We choose candidate images based on following three principles: 
\begin{itemize}
\item The candidate images cannot overlap with the training and validation sets commonly used by image-text retrieval models, preventing data leakage.
\item The candidate images should be in the same domain as the target images to improve the possibility of being similar images.
\item The number of candidate images should be large enough to greatly exceed the size of the original pool, providing a variety of  candidates for similar images.
\end{itemize}

Since both MSCOCO and Flickr30K, the two popular image-text retrieval datasets, contain images from the image sharing website Flickr\footnote{\url{https://www.flickr.com/}}, we adopt other images gathered from Flickr as the additional candidate source for similar images. To be specific, we adopt the 120,000 unlabeled images from COCO\footnote{\url{https://cocodataset.org/}} (unlabeled2017) and combine them with images in the original pool $\mathcal{I_{\textrm{old}}}$ together as candidate images $\mathcal{I_{\textrm{can}}}$.

\begin{figure}[t]
  \centering
  \includegraphics[width=\linewidth]{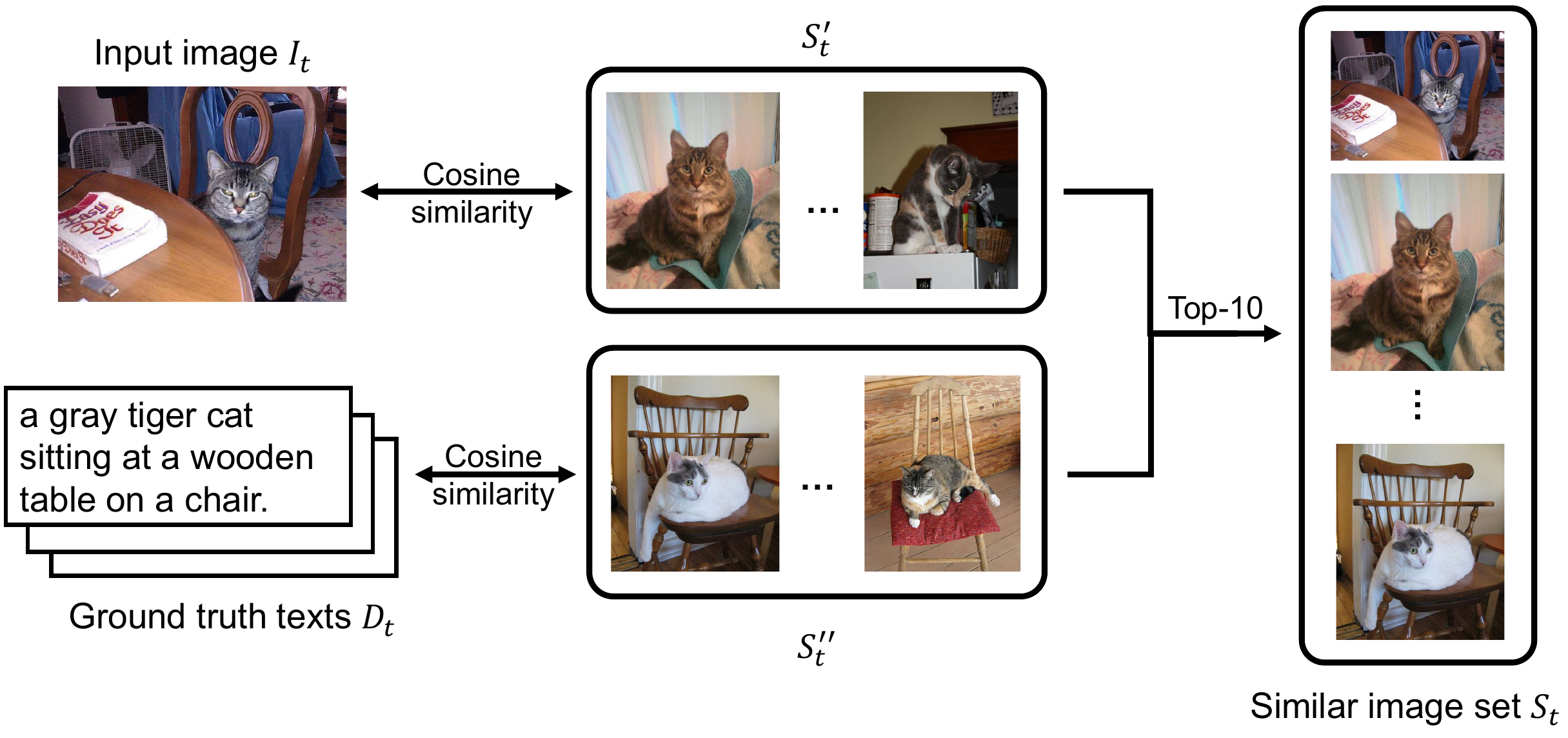}
  \vspace{-16pt}
  \caption{Overview of building a similar image set for a target image via image-to-image and text-to-image matching.}
  \label{fig3}
  \vspace{-8pt}
\end{figure}

\subsubsection{\textbf{Search similar images.}}
Based on candidate images $\mathcal{I_{\textrm{can}}}$, we establish a similar image set $S_t$ for each target image $I_t$ via image-to-image and text-to-image matching, as shown in Figure \ref{fig3}. Specifically, the searching process involves the following three major components:
\begin{enumerate}
\item We first use a vision-language model to extract visual and textual features from images and texts respectively. Concretely, we apply the pretrained CLIP \cite{clip} model to embed all candidate images in $\mathcal{I_{\textrm{can}}}$ including target images in $\mathcal{I_{\textrm{old}}}$, and all texts in $\mathcal{D}$ into a shared semantic space. 
\item Then for each $I_t$ in $\mathcal{I_{\textrm{old}}}$, we compute its cosine similarity scores to all candidate images in $\mathcal{I_{\textrm{can}}}$ and select the Top-$k'$ images as $S_t'$. For its corresponding $M$ text descriptions $D_t=\{d_i\}_{i=1}^M$, we compute their similarity scores to all candidate images as well, and only reserve the Top-$k''$ images as $S_t''$. 
\item Specifically, we select the Top-10 images by combining $S_t'$ and $S_t''$ to form the similar image set $S_t$, which includes the target image $I_t$ and 9 most similar images to $I_t$.
\end{enumerate}

\subsubsection{\textbf{Assemble similar image sets.}}
At last, we assemble all similar image sets $S=\{S_i\}_{i=1}^N$ established from former steps and remove duplicate images from $S$ to obtain the new image pool $\mathcal{I_{\textrm{new}}}$, which contains the entire $\mathcal{I_{\textrm{old}}}$ and large amount of supplemental similar images. The larger pool size and the higher similarity between images in the new pool $\mathcal{I_{\textrm{new}}}$ require stronger fine-grained semantic understanding for image-text retrieval. 

\subsection{Renovate coarse-grained texts}
\label{section:text}
\begin{figure*}[t]
  \centering
  \includegraphics[width=0.95\linewidth]{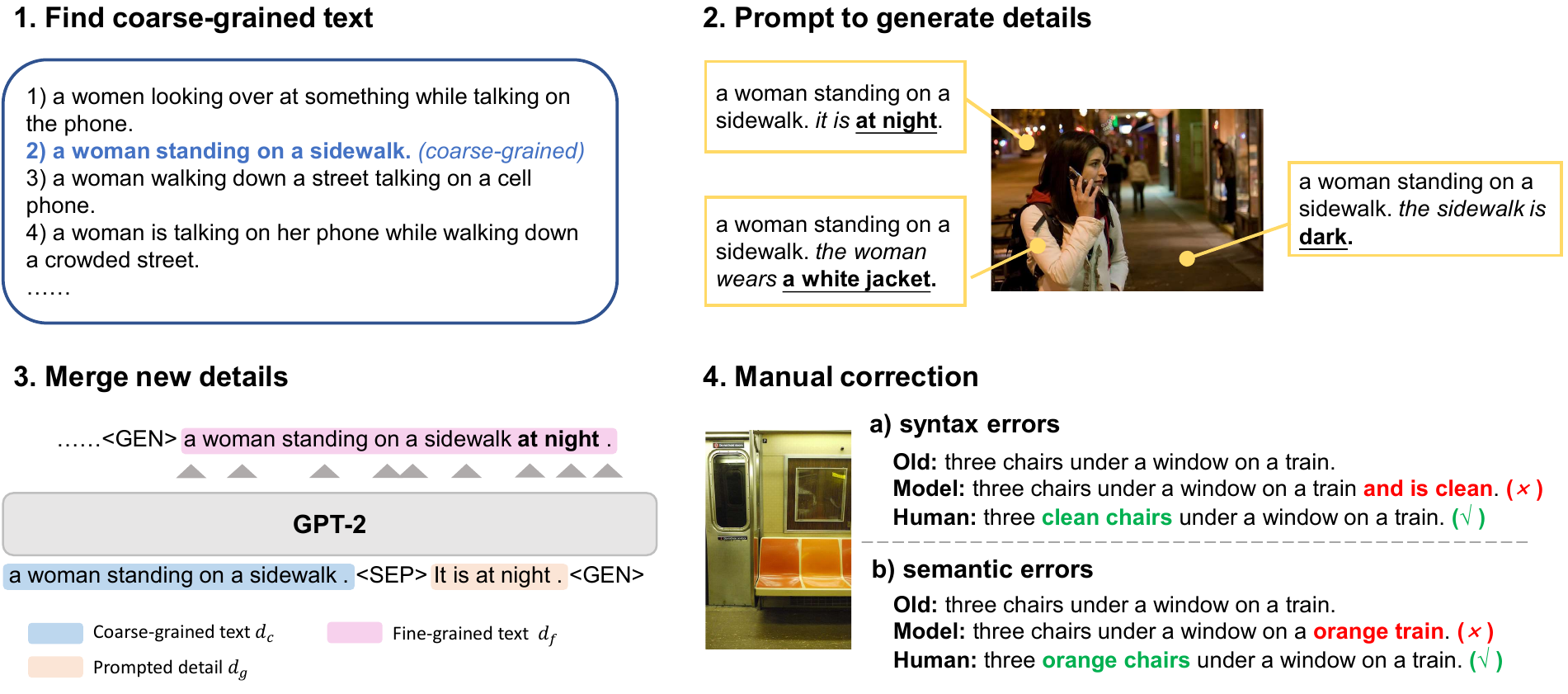}
  \vspace{-8pt}
  \caption{The semi-automatic renovation pipeline for refining the coarse-grained texts into fine-grained ones.}
  \label{fig4}
\end{figure*}

There exists a large number of coarse-grained texts (as illustrated in Figure \ref{fig2}(c)) in existing benchmarks, which could lead to ambiguity problems where different non-target images all match the text. To solve this issue, we propose a semi-automatic renovation pipeline that consists of four steps to harvest fine-grained texts $d_f$ from the coarse-grained texts $d_c$, as shown in Figure \ref{fig4}. 

\subsubsection{\textbf{Find coarse-grained texts.}}
There are both coarse-grained and fine-grained texts in current benchmarks, and only those coarse-grained ones need renovation. 
Since manual checking is expensive and quite time consuming, we leverage the advanced VLP models to perform the coarse-grained text filtering. 
CLIP \cite{clip} has shown competitive performance on various zero-shot image-text tasks including image-text matching. Therefore, we apply a pretrained CLIP in this step. If CLIP is able to rank the target image $I_0$ first given the text description $d_0$, then we consider that $d_0$ is fine-grained enough for the model to distinguish the target image from other images, otherwise we consider $d_0$ as coarse-grained. After such filtering, more than half of the texts in the original benchmark $\mathcal{B}$ are considered as coarse-grained. We then pass them to the following steps to refine them to be finer-grained.

\begin{table}[t]
\caption{Five example prompts to inspire VinVL to generate details from different aspects. X is the placeholder for nouns (objects) appeared in the texts, \emph{e.g.} "sidewalk"}
\vspace{-8pt}
\begin{tabular}{c|l|l}
\toprule
  Row   & Prompts & Details  \\ \midrule
  1  & It is  & new object/place/weather   \\ 
  2  & There is/are  & new object          \\
  3  & The X is/are  & attribute           \\
  4  & The color of X  & color             \\
  5  & The man/woman wears/in  & wearing   \\
\bottomrule
\end{tabular}
\label{table1}
\end{table}

\subsubsection{\textbf{Prompt to generate details.}}
To refine coarse-grained texts filtered from the first step, we need to extract more new details from images and supplement them to the original coarse-grained texts. As the pretrained vision-language models have achieved outstanding performances on image captioning, we adopt a pretrained model to dig more details in images. Particularly, we fine-tune VinVL \cite{vinvl} on the benchmark dataset and use different types of prompts, as shown in Table \ref{table1}, to encourage the VinVL to generate various details based on the original text and image. The detailed process is as follows:

\begin{enumerate}
\item Given a coarse-grained text $d_c$, we first use natural language processing tool Stanford CoreNLP\footnote{\url{https://stanfordnlp.github.io/CoreNLP/}} to analyze and extract $m$ nouns (objects) from $d_c$, which is denoted as $X=\{x_i\}_{i=1}^m$.

\item Create different prompts $P=\{p_i\}_{i=1}^{n'}$ according to the template with nouns in $X$ and separately apply them after $d_c$  to obtain input text as $\{d_c^i\}_{i=1}^{n'}$ where $d_c^i=\langle d_c,p_i\rangle$. 


\item Feed each $d_c^i$ together with corresponding image $I_c$ into VinVL and acquire the new text $d_n=\langle d_c,d_g\rangle$. 

\item Filter the results and only reserve those containing new information. Specifically, we apply clipscore \cite{clipscore} to measure image-text compatibility and discard any $d_n$ whose clipscore is lower than that of $d_c$.
\end{enumerate}

\subsubsection{\textbf{Merge the new details.}}
Note that $d_c$ and $d_n$ are in different formats, where $d_c$ describes the image with only one sentence (\emph{e.g. a woman standing on a sidewalk}), while $d_n$ is in two-sentence format (\emph{e.g. a woman standing on a sidewalk. it is at night}). In order to keep the new text in the same format, we apply a natural language model to merge the two sentences of $d_n$ into one-sentence format (\emph{e.g. a woman standing on a sidewalk at night}). To be specific, we fine-tune GPT-2 \cite{gpt2} to perform sentence merging:

\begin{enumerate}
\item First, we need to construct some training data in the same format as $d_n=\langle d_c,d_g\rangle$ for GPT-2. We first use natural language processing tool Spacy\footnote{\url{https://spacy.io/}} to create a parse tree for each fine-grained text. 
Then we split the content into two parts, corresponding to the detail part $d_g'$ and the rest part $d_c'$ respectively, based on part-of-speech tags and dependency information in the parse tree. 
Specifically, We mainly use two types of structure to construct $d_g'$: \emph{adjective-noun structure} and \emph{Prepositional Phrase}. 
We first extract all pairs of adjective and related object in text, as well as all prepositional phrases. Then we follow the similar format of generated detail sentences $d_g$ to construct $d_g'$ as training data. For example, we split \textit{"a young boy play a frisbee on top of a mountain."} into \textit{"a young boy play a frisbee."} and \textit{"it is on top of a mountain."}

\item We then fine-tune GPT-2 on the training data $d'=\langle d_c',d_g'\rangle$ constructed from above step. Concretely, given the text $\langle d_c',d_g'\rangle$, the model is required to reconstruct $d'$. We hold out some training data to access whether the model learns to merge texts accurately.

\item Feed $d_n$ to the fine-tuned GPT-2 and generate fine-grained text $d_f$ in one-sentence format. As the case shown in step 3 of Figure \ref{fig4}, we feed the original coarse-grained text \textit{"a woman standing on a sidewalk"} and the new detail \textit{"it is at night"} together into GPT-2 and produce a finer-grained text \textit{"a woman standing on a sidewalk at night."}

\item We filter the results again with clipscore and leave out those with scores lower than that of the original text. For cases that have multiple new texts, we choose the best one $d_{best}$ with the highest clipscore as the final renovated text.

\end{enumerate}

\subsubsection{\textbf{Manual correction.}} Even after restrict filtering, some new texts still contain certain errors, usually syntax and semantic errors. We randomly sampled 100 new texts and found that approximately 40\% have some minor errors. Considering that the errors are usually easy to correct, we recruit 20 college students to manually check and correct these errors. For each case with an image and generated text, they are asked to check whether the generated text describes the image correctly, with original text provided to help them quickly locate the newly added details. If there are some errors, they are required to correct them with as few words as possible.
In the end, we accomplish new fine-grained texts by renovating coarse-grained texts through above steps.

\section{Experiments}
\label{section:experiment}

\begin{figure*}[t]
  \centering
  \includegraphics[width=\linewidth]{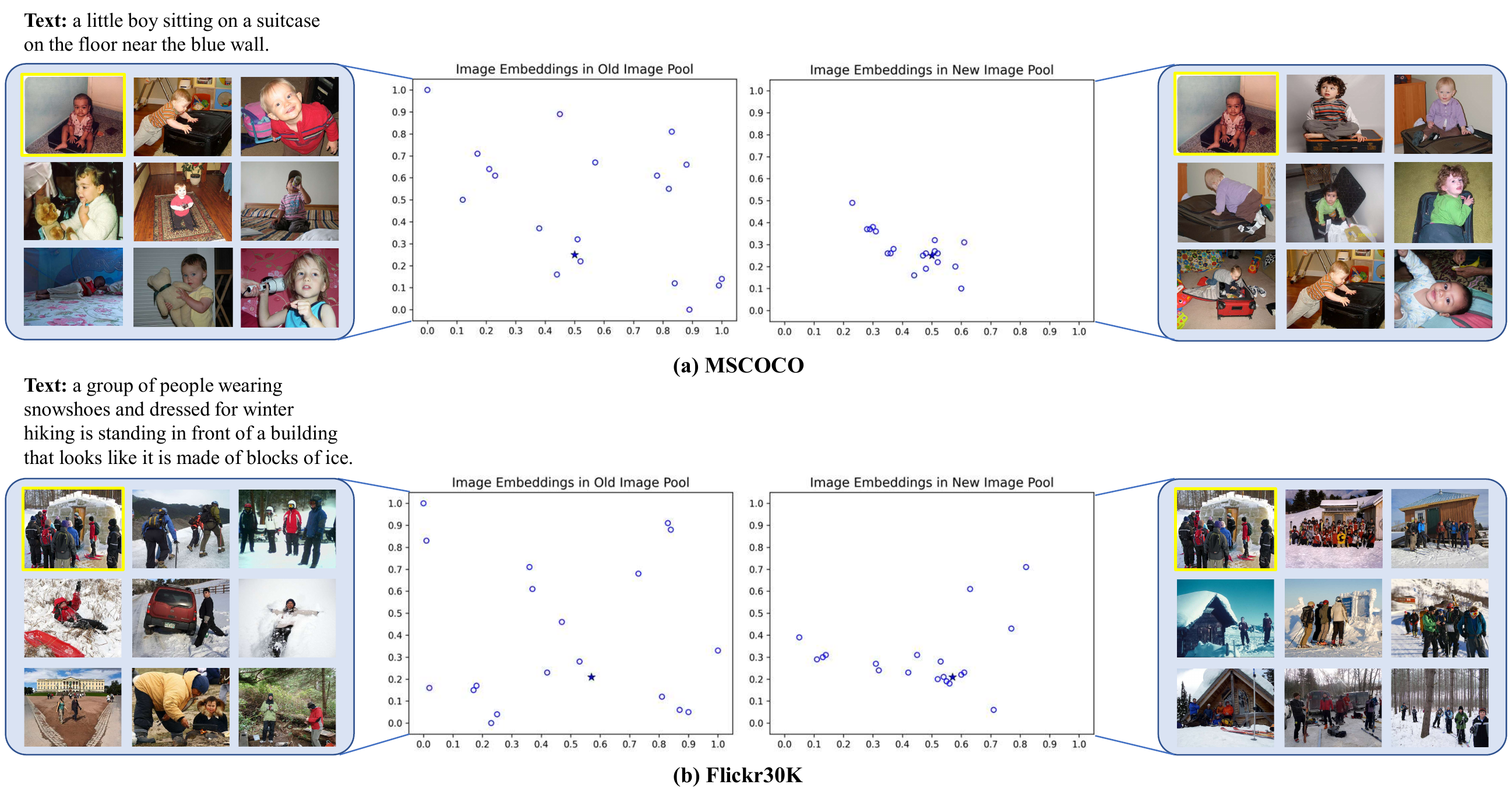}
  \vspace{-16pt}
  \caption{Illustration of the original and our new retrieval image pools on MSCOCO and Flickr30K. We utilize t-SNE to visualize CLIP features of each target image($\star$) and its top-20 most similar images($\circ$) in old pools and new pools respectively. We also present the target image (bordered \textcolor{Gold}{yellow} ) and its top-8 similar images.}
  \label{fig5}
\end{figure*}

\begin{table}[t]
\caption{Size comparison of the original and our new image pool on MSCOCO and Flickr30K.}
\vspace{-8pt}
\begin{tabular}{l|rr}
\toprule
  Image Pool & MSCOCO & Flickr30K  \\ \midrule
  Original   & 5,000  & 1,000   \\
  Our        & \textbf{31,244} & \textbf{6,867}    \\ \bottomrule
\end{tabular}
\label{table2}
\vspace{-8pt}
\end{table}

We renovate the existing benchmarks on MSCOCO and Flickr30K by improving both image pools and texts, and build corresponding new benchmarks named MSCOCO-FG and Flickr30K-FG. In this section, we first present more details about our improved new benchmarks. We then evaluate several classic image-text retrieval models on the new benchmarks. Furthermore, we provide detailed analysis of models in comprehending fine-grained semantics with the new benchmarks.

\subsection{Experiment Settings}
\textbf{Implementation Details.}
If not specified, we use CLIP$_{ViT-B/32}$ as the pretrained model to help searching similar images and filtering coarse-grained texts. We choose VinVL$_{base}$ to generate details with prompts. To merge new details with coarse-grained texts, we adopt GPT-2$_{small}$. We train GPT-2 with learning rate of 3e-4, batch size of 16, and training epochs of 10 on both datasets. We implement our method with PyTorch and conduct all experiments on a single NVIDIA GeForce RTX 1080 Ti GPU.

\noindent \textbf{Evaluation Metrics.}
We follow previous works \cite{vse++,scan,vsrn} to evaluate image-text retrieval performance with Recall@K (R@K), which is defined as the percentage of queries that correctly retrieve the ground-truth in the top-K results. The value of K is 1, 5 and 10.

\subsection{Statistics of New Benchmarks}
The statistics of our new benchmarks and the details of improvements are presented as follows.

\subsubsection{\textbf{Larger and semantically denser image pools}} 
As shown in Table \ref{table2}, the image pools in our new benchmarks are significantly larger than those in the old benchmarks, containing 31k+ and 6k+ images for MSCOCO and Flickr30K respectively. 
To examine the semantic density improvement of images in the new pools, 
we use t-SNE \cite{tsne} to visualize CLIP features of each target image and its top-20 most similar images in the old pools and new pools respectively, in Figure \ref{fig5}. We can see that images in the new pools are semantically closer to each other (or the pool is semantically denser)  than those in the old pools. The top-8 most similar images to the same target image from the old pool and the new pool are illustrated as well. The new pool contains images with more similar objects or scenes to the target image.

\begin{figure*}[t]
  \centering
  \includegraphics[width=0.9\linewidth]{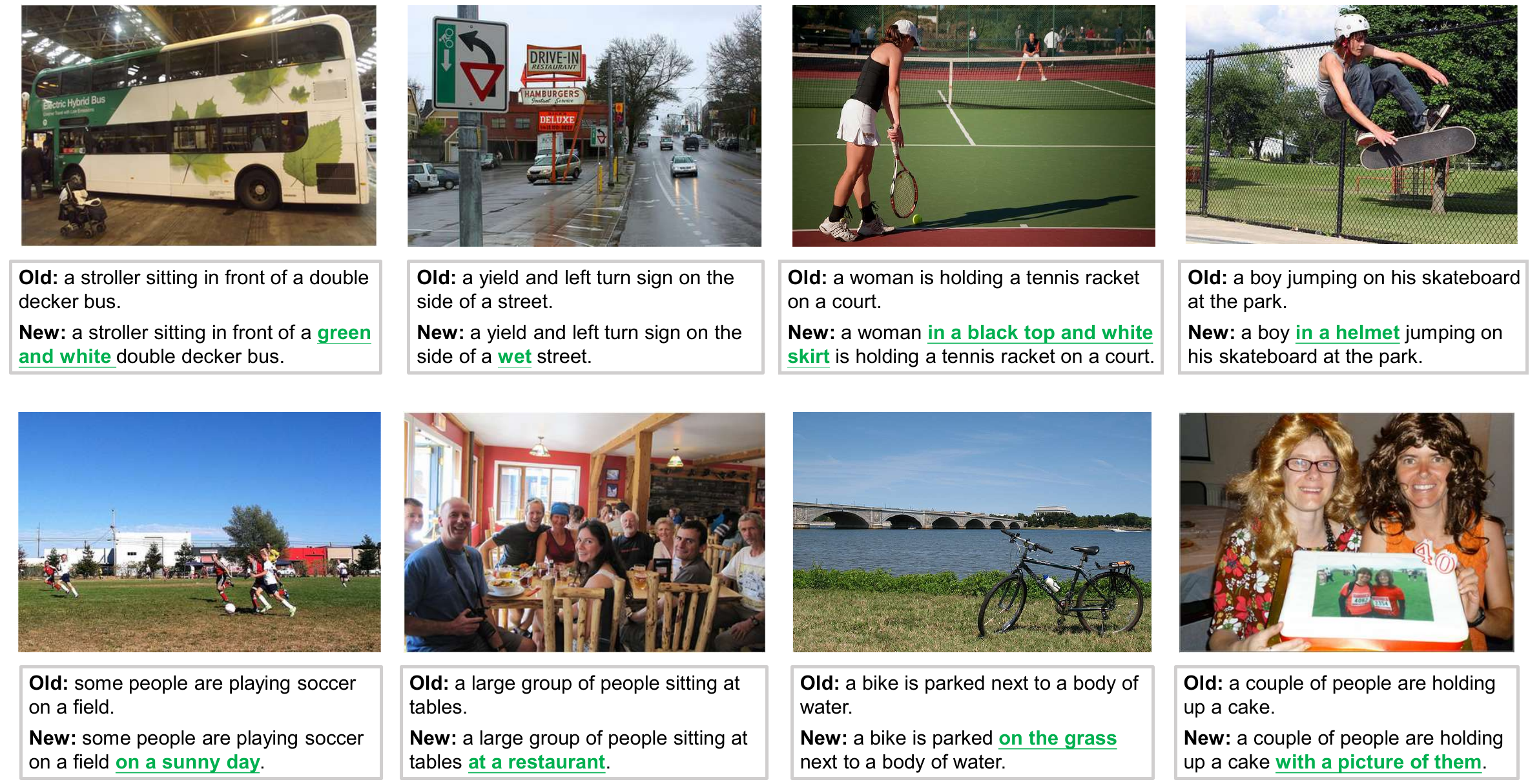}
  \vspace{-8pt}
  \caption{Examples of the old text and improved text for the same image. The \textcolor{Green}{green} words indicate the details that are supplemented through our renovation pipeline.}
  \label{fig6}
  \vspace{-8pt}
\end{figure*}

\subsubsection{\textbf{More detailed and finer-grained texts}} 
We compare the quality of texts before and after renovation from different aspects. We compute the number of nouns and adjectives in all texts, the average number of nouns and adjectives in each sentence, and the average sentence length to measure the granularity of text. The results in Table \ref{table3} show that coarse-grained texts are semantically enriched after renovation on both datasets.
Furthermore, we show some examples of our improved new texts compared to their original coarse-grained texts in Figure \ref{fig6}. Our approach can improve the coarse-grained texts from various aspects, describing more details of the image, including new attributes (\textit{"green and white"}), detailed appearance (\textit{"black top"} and \textit{"white skirt"}), weather (\textit{"sunny day"}), place (\textit{"restaurant"}), and new objects not mentioned (\textit{"a picture"}) etc.

\begin{table}[t]
\caption{Statistics of coarse-grained vs. refined texts.}
\vspace{-8pt}
\subtable[MSCOCO]{
    \begin{tabular}{l|r|rrccc}
    \toprule
    Text Type      & \multicolumn{1}{r|}{\#Texts} & \multicolumn{1}{r}{\#Noun} & \multicolumn{1}{r}{\#Adj} & \multicolumn{1}{c}{\makecell[c]{Avg.\\Noun}} & \multicolumn{1}{l}{\makecell[c]{Avg.\\Adj}} & \multicolumn{1}{l}{\makecell[c]{Avg.\\Length}} \\ \midrule
    Coarse  & \multirow{2}{*}{11,050}   & 38,944 & 6,893 & 3.52 & 0.62 & 10.23  \\
    Refined &    & \textbf{45,075} & \textbf{14,176} & \textbf{4.08} & \textbf{1.28} & \textbf{12.59} \\ \bottomrule
    \end{tabular}
}
\subtable[Flickr30K]{
    \begin{tabular}{l|r|rrccc}
    \toprule
    Text Type      & \multicolumn{1}{r|}{\#Texts} & \multicolumn{1}{r}{\#Noun} & \multicolumn{1}{r}{\#Adj} & \multicolumn{1}{c}{\makecell[c]{Avg.\\Noun}} & \multicolumn{1}{l}{\makecell[c]{Avg.\\Adj}} & \multicolumn{1}{l}{\makecell[c]{Avg.\\Length}} \\ \midrule
    Coarse  & \multirow{2}{*}{2,156} & 7,587 & 1,906 & 3.52 & 0.88 & 10.93 \\
    Refined  &    & \textbf{9,497} & \textbf{2,726} & \textbf{4.40} & \textbf{1.26} & \textbf{13.91} \\ \bottomrule
    \end{tabular}
}
\label{table3}
\vspace{-20pt}
\end{table}

\begin{figure}[t]
  \centering
  \includegraphics[width=\linewidth]{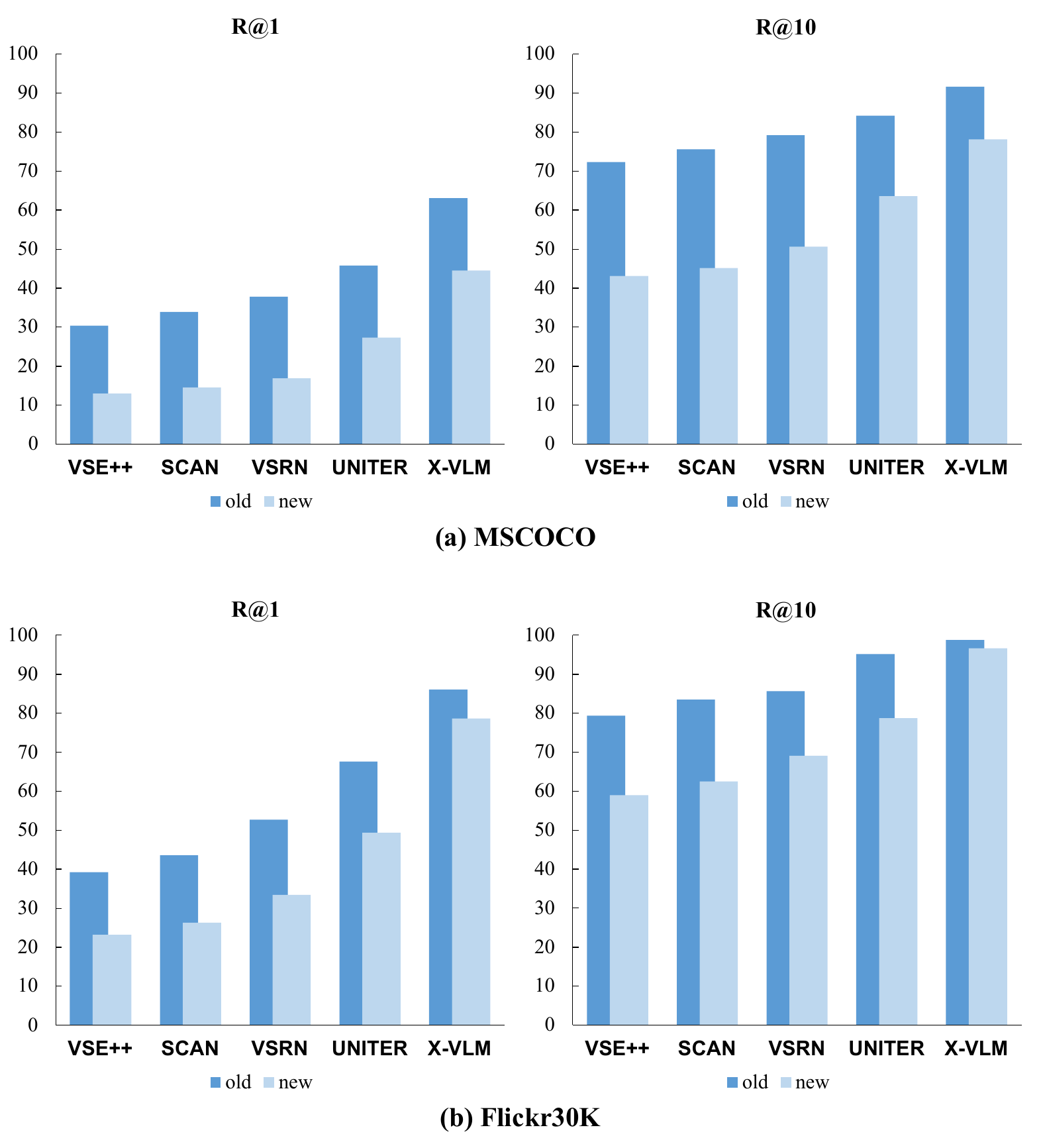}
  \vspace{-20pt}
  \caption{Comparison of different model performances in T2I retrieval on old image pools vs. on new image pools given the new refined text query.}
  \label{fig7}
  \vspace{-4pt}
\end{figure}

\begin{figure}[t]
  \centering
  \includegraphics[width=0.9\linewidth]{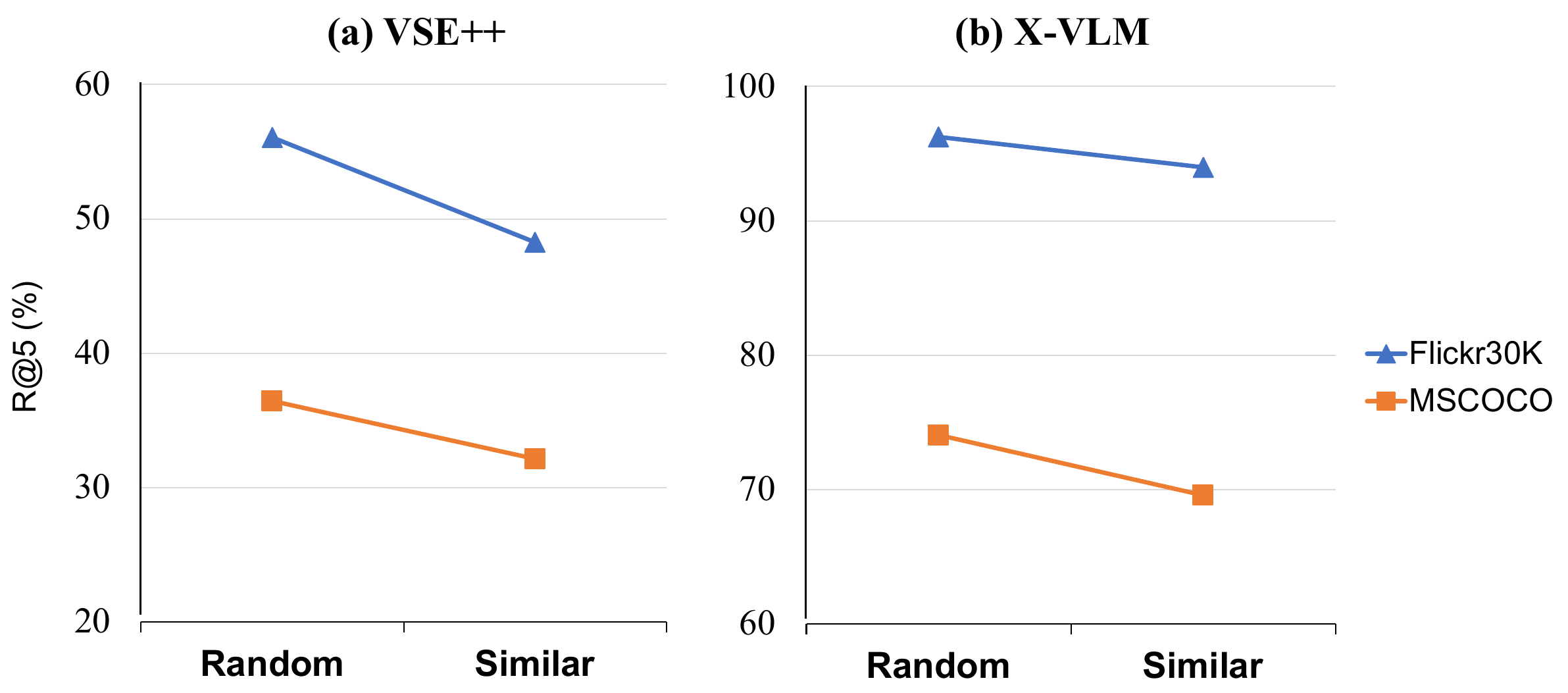}
  \vspace{-8pt}
  \caption{Performance comparison of two models on expanded image pools with random images vs. on our new image pools with similar images.}
  \label{fig8}
  \vspace{-12pt}
\end{figure}

\subsection{Evaluation on New Benchmarks}
We evaluate different existing image-text retrieval models on our new benchmarks to demonstrate the effectiveness of our renovation for making finer-grained evaluations. Specifically, we select three non-pretraining models and two multimodal pretraining models. More details of these models are as follow:

\begin{itemize}
\item \textbf{VSE++} \cite{vse++} encodes images to grid features with CNN and incorporates hard negatives in the loss function.
\item \textbf{SCAN} \cite{scan} first detects objects from images and attends to words in texts to learn latent visual-semantic alignments.
\item \textbf{VSRN} \cite{vsrn} is proposed with a GCN~\cite{gcn} to enhance local visual semantic reasoning and a GRU~\cite{gru} to perform global semantic reasoning.
\item \textbf{UNITER} \cite{uniter} proposes a transformer-based encoder to learn joint visual and textual embeddings with four designed pretraining tasks on four common image-text datasets.
\item \textbf{X-VLM} \cite{xvlm} locates visual concepts in the image given the associated texts and aligns the texts with them by multi-grained vision-language pretraining, which achieves state-of-the-art performances on many Vision-and-Language tasks including image-text retrieval.
\end{itemize}

For non-pretraining models, we train VSE++, SCAN and VSRN on the two benchmark datasets following the same setting. For pretraining models, we fine-tune UNITER$_{base}$ and X-VLM$_{base}$ on the two benchmark datasets.

\subsubsection{\textbf{Experiments on different image pools}}
To validate the effectiveness of enlarging image pools with similar images, we conduct text-to-image (T2I) retrieval experiments on the two benchmarks. The models are required to retrieve the target image from the old image pools or the new image pools given the fine-grained query texts. Figure \ref{fig7} shows the R@1 and R@10 performances of five models on the old pools and the new pools, from which we can see a clear decline of all models when performing retrieval on the new image pools. The performance degradation reveals that our new image pools are more challenging, requiring models to have better fine-grained semantic comprehension ability.

\begin{table}[t]
\caption{Performance R@1 comparison on the new candidate pool with original texts vs. our refined new texts.}
\vspace{-8pt}
\begin{tabular}{l|c|llll}
\toprule
\multirow{2}{*}{Model} & \multirow{2}{*}{\makecell[c]{Refined\\Texts}} & \multicolumn{2}{c}{MSCOCO}                                      & \multicolumn{2}{c}{Flickr30K}  \\
&  & \multicolumn{1}{c}{I2T} & \multicolumn{1}{c}{T2I} & \multicolumn{1}{c}{I2T} & \multicolumn{1}{c}{T2I} \\ \midrule
\multirow{2}{*}{VSE++} & \ding{55} & 41.26 & 10.51 & 52.80 & 19.80 \\
                       & \ding{51} & \textbf{45.44} & \textbf{12.93}  & \textbf{57.20} & \textbf{23.18} \\ \midrule
\multirow{2}{*}{SCAN} & \ding{55} & 45.30 & 11.46 & 68.10 & 22.70 \\
                      & \ding{51} & \textbf{50.90} & \textbf{14.55} & \textbf{72.30} & \textbf{26.28} \\ \midrule
\multirow{2}{*}{VSRN} & \ding{55} & 50.22 & 13.73 & 69.90 & 29.10 \\
                      & \ding{51} & \textbf{55.82} & \textbf{16.83} & \textbf{75.30} & \textbf{33.42} \\ \midrule
\multirow{2}{*}{UNITER} & \ding{55} & 59.08 & 22.05 & 80.70 & 42.86 \\
                       & \ding{51} &  \textbf{64.82} & \textbf{27.32} & \textbf{85.00} & \textbf{49.38}  \\ \midrule
\multirow{2}{*}{X-VLM} & \ding{55} & 80.98 & 38.77 & 96.80 & 73.10 \\
                       & \ding{51} & \textbf{84.16} & \textbf{44.50} & \textbf{97.40} & \textbf{78.70} \\ \bottomrule
\end{tabular}
\label{table4}
\vspace{-8pt}
\end{table}

To further verify the impact of similar images in the new image pools to retrieval evaluation, we build a variant of enlarged image pools. Instead of expanding with similar images, we expand the image pool with randomly selected images from the same candidate image source to the same pool size. We then conduct T2I comparison experiments on the randomly expanded image pools and our new pools. Figure \ref{fig8} shows the R@5 performances of VSE++ and X-VLM on these two types of image pools. We observe much lower performances on similar image pools for the same model, which indicates that the introduction of similar images to image pools indeed increases the retrieval difficulty, which requires the retrieval model to have better cross-modal fine-grained semantic understanding and matching abilities.

\subsubsection{\textbf{Experiments on texts with different granularity}}
We also carry out text-to-image (T2I) retrieval and image-to-text (I2T) retrieval experiments to verify the effectiveness of our improved texts.  We compare the retrieval performance of the same model on the new candidate pool with the original texts or our new refined texts. The results in Table \ref{table4} show that models all perform better on R@1 when given the new renovated texts no matter as queries or as candidates, which demonstrates that the improved texts are finer-grained, leading to better cross-modal matching. 


\subsection{Empirical Study of Classic Models}
Above experiments show that our new benchmarks can better evaluate the model capability in fine-grained semantic comprehension and alignment. We then try to provide more analysis and insight into the question we raise up at the beginning through experiments on our new benchmarks: do current models really have human-comparable cross-modal semantic knowledge, especially in fine-grained semantic understanding?

As we can see in Figure \ref{fig7}, all models achieve much lower recall for T2I retrieval on our new benchmarks, in which even state-of-the-art model X-VLM shows a significant degradation, especially on MSCOCO-FG, achieving only 44.5 on R@1. These results demonstrate that \textbf{current image-text retrieval models are still far away from perfect in fine-grained cross-modal semantic understanding.}

Moreover, we notice that different models exhibit different degrees of degradation on the new benchmarks. Pretraining models (UNITER, X-VLM) still perform much better than non-pretraining models (VSE++, SCAN, VSRN) on MSCOCO-FG and Flickr30K-FG, demonstrating the advantage of pretraining models in gaining more cross-modal semantic knowledge by seeing a great deal of image-text pairs and through the pretraining process with specific alignment tasks. 
Different non-pretraining models and pretraining models also perform differently. Non-pretraining models equipped with additional modules for fine-grained semantic comprehension and alignment like VSRN perform much better than those without (\emph{i.e.} VSE++, SCAN), while pretraining models trained with specific fine-grained pretraining tasks like X-VLM tend to perform better than those without (\emph{i.e.} UNITER).

\begin{figure*}[t]
  \centering
 \includegraphics[width=0.87\linewidth]{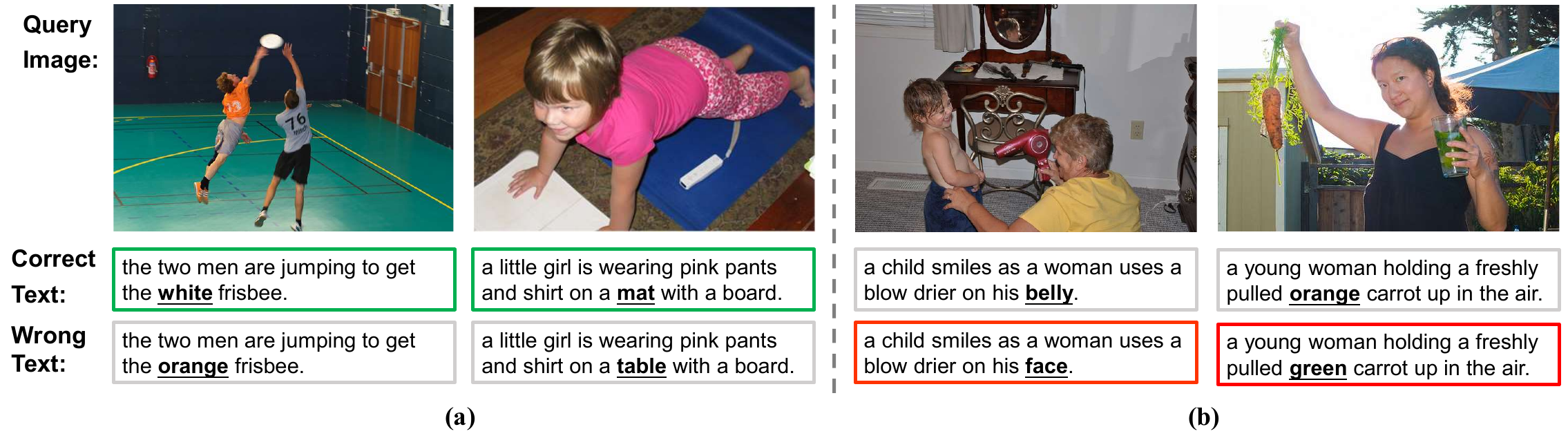}
  \vspace{-12pt}
  \caption{Examples of X-VLM succeed or fail in choosing the correct text for a given image.}
  \label{fig10}
  \vspace{-8pt}
\end{figure*}

\begin{figure}[t]
  \centering
  \includegraphics[width=0.73\linewidth]{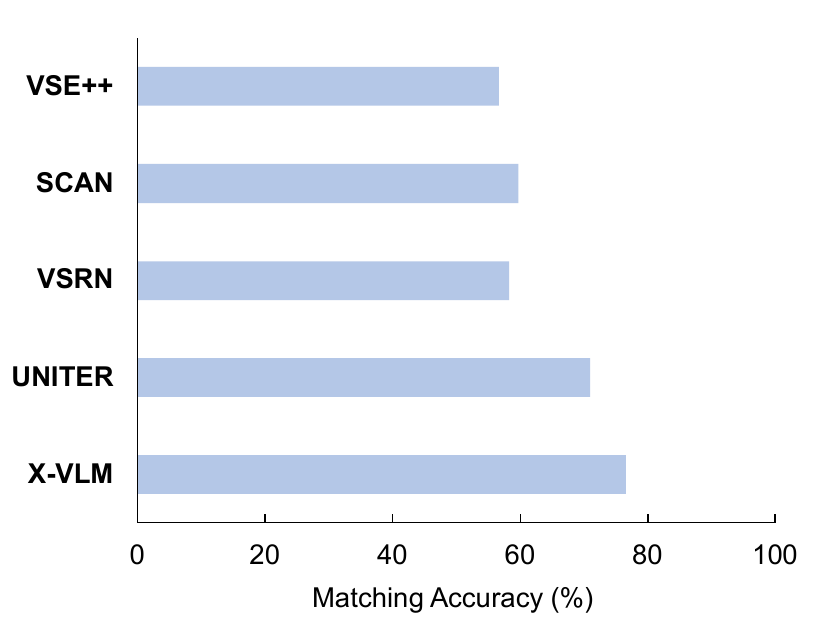}
  \vspace{-12pt}
  \caption{Comparison of different model performances in matching image with wrong/correct text pairs.}
  \label{fig9}
  \vspace{-16pt}
\end{figure}

Although current models have made some effective efforts to improve fine-grained semantic understanding between image and text, there is still much room for improvement.
We further carry out some additional experiments to inspire new ideas for future directions in fine-grained semantic understanding. 
Specifically, we utilize the wrong texts generated by models in the semi-automatic renovation pipeline and their corresponding human-corrected texts in MSCOCO-FG to further study the capability of models in fine-grained semantic understanding. According to our observation in the correction process, the wrong texts are almost the same as the correct texts, except for a few words. Given an image and a pair of wrong text and correct text, the model is required to choose the best one for the image, which can examine the capability of model in distinguishing and aligning details across two modalities. 
The results in Figure \ref{fig9} confirm that pretraining models perform better than non-pretraining models on fine-grained semantic understanding, while they still have some difficulty in choosing correct texts as their matching accuracy is still below 80\%. 

We further study the good and bad matching cases of X-VLM and show some examples in Figure \ref{fig10}. The advanced image-text retrieval models like X-VLM have demonstrated good ability in understanding and distinguishing objects and their attributes at a fine-grained semantic level, as shown in Figure \ref{fig10}(a). However, it still struggles to distinguish the attributes of adjacent objects in images, as shown in Figure \ref{fig10}(b), which suggests that future works of image-text retrieval can pay more attention to improving model understanding of fine-grained object-object relation and object-attribute alignment. For example, consider training models with similar images and confusing texts like above as hard negative samples to gain better fine-grained semantic understanding ability.


\vspace{-4pt}
\section{Conclusions}
In this work, we review the common benchmarks for image-text retrieval and identify that they are insufficient to evaluate the true capability of fine-grained cross-modal semantic alignment, due to the coarse-grained images and texts. We therefore establish the improved new benchmarks called MSCOCO-FG and Flickr30K-FG based on the old benchmarks. Specifically, we enlarge the original image pool by adopting more similar images and propose a semi-automatic renovation pipeline to upgrade coarse-grained texts into finer-grained ones with little human effort. We demonstrate the effectiveness of our renovation by evaluating representative image-text retrieval models on our new benchmarks and analyze the model capabilities on fine-grained semantic comprehension through extensive experiments. The results show that there is still much room for improvement in fine-grained semantic understanding, even for state-of-the-art models, especially in distinguishing attributes and relations of close objects in images.


\textcolor{black}{Note that our renovation process can leverage different pretraining language models and vision-language models to improve the benchmarks with ensembling. 
Due to the limitations of the vision-language model in prompting new details, we only provide the base version with specific models in this work, which might introduce bias in selecting images and renovating texts. 
As the large pretraining models advance, we can further benefit from them and improve our benchmarks accordingly. We hope the new benchmarks will inspire further in-depth research exploration on cross-modal retrieval.
}

\vspace{-16pt}
\section*{Acknowledgements}
This work was partially supported by National Key R\&D Program of China (No.2020AAA0108600) and National Natural Science Foundation of China (No.62072462).

\bibliographystyle{ACM-Reference-Format}
\bibliography{7_Reference}


\end{document}